\pdfoutput=1

\documentclass[11pt]{article}
\PassOptionsToPackage{sort&compress}{natbib}
\usepackage[hyperref]{acl}
\usepackage[utf8]{inputenc}
\usepackage{times}
\usepackage{latexsym}
\usepackage[T1]{fontenc}
\usepackage{amsmath,amsfonts,amstext,bm}
\usepackage{microtype}
\usepackage{inconsolata}
\usepackage{setspace}
\usepackage[capitalise,nameinlink,noabbrev]{cleveref}
\usepackage{adjustbox}
\usepackage{graphicx}
\usepackage{booktabs}
\usepackage{siunitx}
\usepackage{pseudocode}
\usepackage{tabularx}
\usepackage{multirow}
\usepackage{pifont}
\usepackage[normalem]{ulem}
\usepackage{import}

\usepackage{enumitem}
\setlist{itemsep=0pt,parsep=2pt}

\newcommand{\cmark}{\ding{51}}
\newcommand{\xmark}{\ding{55}}

\makeatletter
\patchcmd{\@@setcref}         {??}{\color{orange} ??}{}{}
\patchcmd{\@@setcref}         {??}{\color{orange} ??}{}{}
\patchcmd{\@@setcrefrange}    {??}{\color{orange} ??}{}{}
\patchcmd{\@@setcrefrange}    {??}{\color{orange} ??}{}{}
\patchcmd{\@@setcrefrange}    {??}{\color{orange} ??}{}{}
\patchcmd{\@@setcrefrange}    {??}{\color{orange} ??}{}{}
\patchcmd{\@@setcrefrange}    {??}{\color{orange} ??}{}{}
\patchcmd{\@@setcrefrange}    {??}{\color{orange} ??}{}{}
\patchcmd{\@@setnamecref}     {??}{\color{orange} ??}{}{}
\patchcmd{\@@setnamecref}     {??}{\color{orange} ??}{}{}
\patchcmd{\@@setcpageref}     {??}{\color{orange} ??}{}{}
\patchcmd{\@@setcpageref}     {??}{\color{orange} ??}{}{}
\patchcmd{\@@setcpagerefrange}{??}{\color{orange} ??}{}{}
\patchcmd{\@@setcpagerefrange}{??}{\color{orange} ??}{}{}
\patchcmd{\@@setcpagerefrange}{??}{\color{orange} ??}{}{}
\patchcmd{\@@setcpagerefrange}{??}{\color{orange} ??}{}{}
\patchcmd{\@@setcpagerefrange}{??}{\color{orange} ??}{}{}
\patchcmd{\@@cref}            {??}{\color{orange} ??}{}{}
\makeatother

\title{Voices in a Crowd: Searching for Clusters of Unique Perspectives}

\author{Nikolas Vitsakis\textsuperscript{\textdagger} \\
  Heriot-Watt University \\
  \texttt{nv2006@hw.ac.uk} \\\And
  Amit Parekh\textsuperscript{\textdagger} \\
  Heriot-Watt University \\
  \texttt{amit.parekh@hw.ac.uk} \\\And
  Ioannis Konstas \\
  Heriot-Watt University \\
  \texttt{i.konstas@hw.ac.uk}
  }

\begin{document}
\maketitle

\begingroup\def\thefootnote{\textsuperscript{\textdagger}}\footnotetext{Equal contribution}\endgroup

\begin{abstract}
Language models have been shown to reproduce underlying biases existing in their training data, which is the majority perspective by default. Proposed solutions aim to capture minority perspectives by either modelling annotator disagreements or grouping annotators based on shared metadata, both of which face significant challenges. We propose a framework that trains models without encoding annotator metadata, extracts latent embeddings informed by annotator behaviour, and creates clusters of similar opinions, that we refer to as \textit{voices}. Resulting clusters are validated post-hoc via internal and external quantitative metrics, as well a qualitative analysis to identify the type of voice that each cluster represents. Our results demonstrate the strong generalisation capability of our framework, indicated by resulting clusters being adequately robust, while also capturing minority perspectives based on different demographic factors throughout two distinct datasets.%
\footnote{All code is made available at \url{https://github.com/Ni-Vi/Cluster}.}

\textcolor{red}{\textbf{Content Warning:} This document contains and discusses examples of potentially offensive and toxic language.}

\end{abstract}

\begin{figure*}[!ht]
    \centering
    \includegraphics[width=\textwidth]{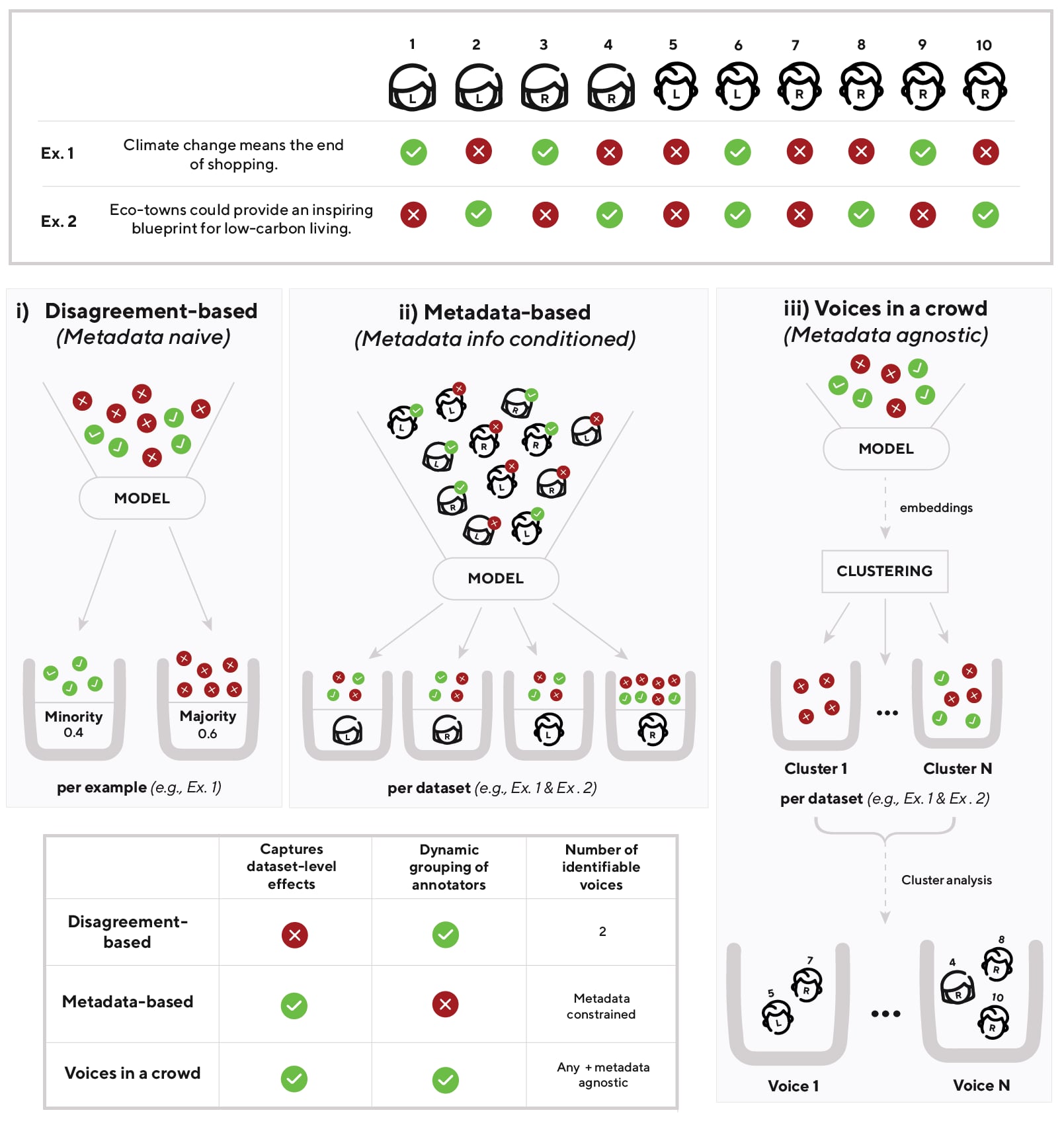}
    \caption{Different approaches for handling annotations: i) disagreement-based create per-example distributional labels which fail to account for dataset-level effects; ii) metadata-based train models on annotations linked with annotator metadata, which often groups disagreeing annotators who share metadata labels; iii) the \textit{``Voices in a crowd''} approach dynamically creates clusters based on annotation patterns and finally verifies each cluster as a voice based on post-hoc matched metadata labels.}
    \label{fig:figure1}
\end{figure*}

\section{Introduction}

Supervised training is rooted in the presupposition that every example in a dataset has a single ground truth, also known as the gold label \citep{hettiachchi2021challenge}. However, disagreement among dataset annotators challenges the notion that a single, per-example, ground truth exists \citep{uma2022scaling, uma2021learning}. While disagreement can be indicative of task difficulty or semantic ambiguity \citep{wang2021deep, jiang2022investigating, sandri2023don}, it can also indicate the existence of both stable \textit{and} conflicting inter-annotator perspectives \citep{basile2020s, abercrombie2023consistency}. 

Nevertheless, capturing minority perspectives present in the data, which we parallel to \textit{voices in a crowd}, has proven challenging. Two main approaches attempt to move beyond gold labels: \textbf{i) disagreement-based} which leverage annotator disagreement to provide distributional per-item prediction labels \citep{leonardelli2023semeval, uma2022scaling, uma2021learning}, and \textbf{ii) metadata-based}, which encode annotator metadata to boost the signal from voices with the same metadata labels \citep{fleisig2023majority, gupta2023same, beck2023not} (i and ii respectively in \cref{fig:figure1}). 

However, both approaches come with strong vulnerabilities. \textbf{Disagreement-based} approaches collapse multiple minority voices into a singular, per-item, minority-majority distribution \citep{gordon2022jury}, essentially limiting the number of expressed voices to the number of predicted labels (i.e., two voices in a binary prediction task). On the other hand, while \textbf{metadata-based} approaches allow for multiple minority voices to be expressed (albeit limited by metadata collected), they are based on the erroneous assumption that most members that share metadata labels (e.g., gendered females) will also exhibit similar patterns of behaviour \citep{2023arXiv230907034B, dang2020self}. 

We introduce a framework that addresses both issues (\cref{fig:figure1}iii): it forms multiple clusters of distinct voices \textit{solely} based on \textit{annotator behaviours exhibited during the annotation process} in an unsupervised manner. Our pipeline trains models to predict \textit{each annotation made by each annotator} for a given text input.

The final hidden states form what we refer to as \textit{behavioural embeddings}, representing how a given annotator will behave when shown that text sample, are then clustered via unsupervised methods. We define each created cluster 
as a potential \textit{voice}---a group perspective of annotators with similar annotating behaviours. 

We apply our framework to two datasets related to political bias that have been found to contain multiple heterogeneous and conflicting perspectives \citep{de2020measuring, menini2016agreement, nemeth2023scoping, chen2019seeing}. To identify the group whose voice each cluster belongs to, we match each data point with annotator metadata post-hoc while we also conduct an in-depth qualitative analysis of the clusters themselves. The resulting clusters show high internal label consistency of either \textbf{i) dataset majority labels} (e.g., left-leaning in a left-leaning majority dataset), \textbf{ii) dataset minority labels} (e.g., right-leaning in a left-leaning majority dataset), but most importantly their intersection resulting in \textbf{iii) inter-minority labels} (e.g., right-leaning and highly educated, in a left-leaning, non-highly educated majority dataset). We are the first to dynamically identify voices of minority opinions within larger majority/minority groups, highlighting the significance of providing an intersectional understanding of annotators that goes beyond current grouping methodologies.

\section{Related Work}\label{sec:related-work}

\paragraph{Disagreement-Based Solutions} As an alternative to gold labels, recent research has introduced the use of silver labels, i.e., distributional per-item labels that measure disagreement amongst annotators \citep{leonardelli2023semeval,uma2022scaling,uma2021learning}. While such approaches allow for the identification of controversial examples in datasets \citep{fornaciari-etal-2022-hard}, they fail to capture stable inter-annotator disagreements throughout the dataset that could provide insight as to why disagreement occurs beyond an item-by-item scale \citep{abercrombie2023consistency, vitsakis2023ilab}. 

To be more specific, disagreement-based solutions essentially limit the number of possible expressed voices into the number of predicted labels; the upper bound of possible voices expressed in a binary task is always two, no matter how diverse the dataset. Unfortunately, this type of aggregation leads to the erasure of what we define as \textit{inter-minority voices}: stable opinions held by minority groups that are in conflict with each other as well as the majority, across examples.

\paragraph{Metadata-Based Solutions} A recent trend aiming to capture diverse perspectives has attempted to group annotators based on
their metadata. Such approaches encode collected annotator metadata, such as annotator beliefs \citep{rottger2021two, davani2023hate} or demographics  \citep{fleisig2023majority, gupta2023same}, into the training pipeline to allow learning of patterns between annotations and in-group tendencies. While the incorporation of such information can seemingly improve model performance in specific tasks \citep{welch-etal-2020-compositional}, evidence suggests that such results might be dataset-specific \citep{lee-etal-2023-large}.

This is due to the assumption that annotators sharing metadata labels will behave similarly during the annotation process. However, demographics are not necessarily predictive of underlying behaviour \citep{hwang-etal-2023-aligning, 2023arXiv230907034B}, while social sciences have also explained similar issues with self-reported measures \citep{dang2020self, schwarz1999self}. With the added issue that annotator metadata is often not collected outright \citep{prabhakaran2021releasing}, there is a direct need for methodologies that identify distinct group voices based on factors other than a-priori collected labels. 

\paragraph{Unsupervised Learning and Clusters of Voices} To circumvent previously mentioned issues, unsupervised learning could be employed along the lines of how past research identified emergent themes within corpora via clustering of latent textual embeddings \citep{sevillano2007text, meng2022topic, dhillon2001concept}. Recently, \citet{meng2022topic} showed promising results in automatic topic discovery by utilising pretrained language models to cluster representations in a \textit{joint latent space}: formed by combining latent spaces of multiple modalities during learning, in this case word and document level embeddings. We aim to take this work further through our use of joint behavioural embeddings, informed by both text and annotating behaviour, to automatically find \textit{voices}, i.e., clusters of similar opinions.

There are significant challenges to this approach. Fine-tuning pretrained language model embeddings produce embeddings that are often anisotropic and anisometric \citep{rajaee2021does,xu2021cross}; when paired with their high dimensionality nature, clustering via distance-based metrics is challenging. By using appropriate dimensionality reductions \citep{mu2017all, cai2020isotropy}, the relationships between features can be analysed and clustered through Euclidean distance-based metrics \citep{mcinnes2018umap}.

\section{Experimental Setup}

Our framework comprises a \textbf{supervised} and an \textbf{unsupervised component}. The former
produces latent embeddings informed by both text and annotating behaviour that the latter uses to cluster into voices. Being the first such approach, we compared performance across a variety of transformer-based architectures, clustering, and dimensionality reduction techniques to identify optimal combinations. 

The \textbf{supervised component} explores several modelling choices (\cref{sec:training-models}) fine-tuned on each dataset to predict each annotator's individual annotation for a given example without providing any annotator metadata that could bias the model \citep{vitsakis2023ilab}. The \textbf{unsupervised component} then performs dimensionality reduction on the \textit{behavioural embeddings}---the final hidden states from the supervised component---and finally creates clusters via several unsupervised algorithms (\cref{sec:dimensionality_reduction_techniques}). 
Clusters are evaluated through internal (i.e., intra-cluster similarity) and external metrics (i.e., consistency of demographic labels in a given cluster), and via qualitative analysis of the best-performing combination of components.

\subsection{Datasets}\label{sec:datasets}

All datasets used in our experiments contain the following annotator demographics: personal political leaning, age, and education level.

\textbf{Media Bias Annotation Dataset} \citep[MBIC;][]{spinde2021automated,spinde2021mbic} comprises sentences from media articles that may contain political bias from news outlets across the political spectrum (e.g., Fox News, MSNBC, etc.) covering 14 potentially divisive topics (e.g., gender issues, coronavirus, the 2020 American election). 784 crowd-sourced annotators labelled sentences on whether they consider them to contain bias. Demographics were slightly skewed in political ideology (44.3\% left-leaning, 26.7\% right-leaning, 29.1\% center).

\textbf{Global Warming Stance Dataset} \citep[GWSD;][]{luo2020detecting} contains opinions of varying intensities on the subject of global warming, gathered from news outlets with different political leanings (e.g., The New York Times, Breitbart). 398 annotators labelled each sentence with whether they agreed, disagreed, or were neutral. Demographic skew of this dataset mirrored that of MBIC in self-reported political affiliation (46\% Democrat, 21.2\% Republican, 28.8\% Independent, 4\% Other).

\section{Supervised Component}\label{sec:training-models}

Each of the following modelling architectures was trained through a different combination of inputs (visual representation in \cref{sec:model_image}): 
given a text sample in a dataset, $\mathbf{x} \in \mathbf{X}$, we predict the \textit{individual annotation} of each annotator $p_\theta(\mathbf{y} | \mathbf{x})$ where $\mathbf{y} = (y_1, \dots, y_K)$ and $K$ is the total number of unique annotators within the dataset.

\noindent \textbf{Unpooled Cross Attention} uses a pretrained T5 encoder \citep{raffel2020exploring} where the encoded text and the embedded annotator unique identifiers are fed through a decoder to predict each annotator's annotation as a sequence. Annotator embeddings are directly informed by the text via a cross-attention layer aiming to capture the influence of the text in the annotators' behaviours. 

\noindent \textbf{Pooled Cross Attention} follows \citet{sullivan2023university}, which showed strong performance in predicting annotator disagreement in the 2023 Learning With Disagreements shared task \citep[LeWiDi;][]{leonardelli2023semeval}. This model is similar in structure to \textit{Unpooled Cross Attention} since it also uses a T5 encoder as the backbone. However, the dimension for each encoded text token is downsampled, as previous research has indicated possible benefits in the salience of encoded features \citep{schick2019attentive, dhingra2018embedding, holzenberger2018learning}. Finally, decoder outputs are pooled \citep{reimers2019sentence} to predict an aggregated annotation for each batch.

\noindent \textbf{Encoder-Encoder} treats text and annotators as separate modalities, inspired by vision and language models \citep{tan2019lxmert, singh2022flava, agarwal2020history}.
The encoded text (using T5) and embedded annotator IDs are concatenated and fed through a bidirectional encoder 
to predict the annotation of each annotator, allowing for interaction between text and annotator embeddings. 

\noindent \textbf{Classifier Model} simply concatenates the text with the unique annotator identifier, before passing to an encoder (BERT; \citealp{DBLP:journals/corr/abs-1810-04805} for GWSD, and RoBERTa; \citealp{DBLP:journals/corr/abs-1907-11692} for MBIC) to predict each annotation label independently. The independence between annotators limits interaction between annotators during training.

\begin{table}[]
\centering
\small
\sisetup{mode=match,tight-spacing=true,separate-uncertainty=true,table-align-uncertainty=true,group-separator=\pm,detect-weight=true}
\renewcommand{\arraystretch}{1.3}
\begin{tabularx}{\linewidth}{X S[table-format=1.2] S[table-format=1.2(2.2)] @{}}
    \toprule
    & {F1 Score\;$\uparrow$} & {APCS\;$\downarrow$}  \\
    \midrule
    
    \multicolumn{1}{@{}l}{\textit{GWSD Dataset}} \\

    Unpooled Cross Attention & \textbf{0.65} & \bfseries 0.14 (0.07) \\ 
    Pooled Cross Attention & 0.19 & 0.54 (0.13) \\ 
    Encoder-Encoder & 0.63 & 0.15 (0.11)\\ 
    Classifier Model & 0.63 & 0.81 (0.14)\\
    Pretrained Decoder & 0.62 & 0.66 (0.08)\\
    Pretrained Encoder-Decoder & 0.19 & 0.95 (0.02) \\
    
    \multicolumn{1}{@{}l}{\textit{MBIC Dataset}} \\

    Unpooled Cross Attention & \textbf{0.72} &  0.22 (0.05)\\ 
    Pooled Cross Attention & 0.43 & 0.70 (0.06)\\ 
    Encoder-Encoder & \textbf{0.72} & \bfseries 0.21 (0.06) \\ 
    Classifier & 0.38 & 1.00 (0.00)\\
    Pretrained Decoder  & 0.63 & 0.75 (0.07)\\
    Pretrained Encoder-Decoder & 0.71 & 0.74 (0.25) \\
    
    \bottomrule
    
\end{tabularx}

\caption{Overall performance (F1 Score) for the supervised component of our framework (6 modelling architectures) on MBIC and GWSD for the task of individual annotator prediction. We also report the Average Pairwise Cosine Similarity (APCS) across the final hidden states; a lower score indicates greater variety in representation which correlates with better clustering performance.}
\label{tab:model_performance}
\end{table}

\noindent \textbf{Pretrained Decoder} 
is a decoder-only GPT-2 model \citep{radford2019language} prompted with the concatenated text and annotator identifiers in the form  \mbox{``\texttt{<text> [SEP] <Ann 1> [SEP] ... <Ann K>}''} and predicts the annotation for each annotator. 

\noindent \textbf{Pretrained Encoder-Decoder} is similar to \textit{Unpooled Cross Attention}. It uses a pretrained T5 encoder-decoder instead; the only difference is that the unique annotator identifiers are embedded through the decoder of the T5 model itself---instead of a decoder trained from scratch---to predict each annotator's annotation autoregressively. The decoder is unidirectional, forcing casual attention between annotators in their canonical order. It is limited compared to the \textit{Encoder-Encoder}, despite using cross-attention. 

\paragraph{Metrics} We compute F1 score to measure the accuracy of predictions, and Average Pairwise Cosine Similarity (APCS) between hidden states of predicted annotations to illustrate how dense the latent states are by the end of training; we show that lower scores generally correlate with better clustering performance (see \cref{sec:quantitative_validation}).

\paragraph{Results} 
\cref{tab:model_performance} summarises the results. For GWSD, Unpooled Cross Attention achieved the highest F1 score and lowest APCS, whereas it shared a similar performance with Encoder-Encoder for MBIC (albeit the latter has slightly lower APCS). This could be down to the bidirectional attention mechanism (either through cross-attention or encoder self-attention) between the annotator embeddings and the text during training. 

These results also showcase the importance of reporting on the quality of the hidden states. For example, while the Pretrained Encoder-Decoder and Classifier Model have high F1 scores on the MBIC and GWSD datasets respectively, their low scores on APCS indicate dense hidden states that would result in poor clustering outcomes. Overall, our findings show that the bidirectional attention-based models that allow interaction between text and annotator embeddings are the only consistent architectures to show high F1 and low APCS scores.

\begin{table*}[t]
\centering
\footnotesize
\sisetup{mode=match,tight-spacing=true,separate-uncertainty=true,table-align-uncertainty=true,group-separator=\pm,detect-weight=true,detect-inline-weight=math}
\renewcommand{\arraystretch}{1.3}
\begin{tabularx}{\textwidth}{@{} X S[table-format=2] *{4}{S[table-format=1.2]} *{2}{S[table-format=2.1]} @{}}
    \toprule
    &&&&
    \multicolumn{2}{c}{\textit{Purity}\;$\uparrow$} &
    \multicolumn{2}{c}{\textit{Prototypical Cluster \%}\;$\uparrow$}
    \\
    \cmidrule(lr){5-6}
    \cmidrule(lr){7-8}
    & {\# Clusters} 
    & {DB Index\;$\downarrow$} 
    & {Silhouette\;$\uparrow$} 
    & {Political} 
    & {Education} 
    & {Political} 
    & {Education} 
    \\
    \midrule
    \textit{Unpooled Cross Attention} \\
    \quad No dim. reduction & 19 & 6.35 &  0.02 & \textbf{0.71} & \textbf{0.71} & 15.8 & 0.0  \\
    \quad w/ PCA  & 10 & 1.98  &  0.10 & 0.36 & 0.43 & 20.0 & 0.0  \\
    \quad w/ UMAP & 19 & 0.81  &  0.47  & 0.38 & 0.42 & 31.6 & 5.3  \\
    \textit{Pooled Cross Attention} \\ 
    \quad No dim. reduction & 19 & 3.03  &  0.06  & 0.42 & 0.48 & 26.0 & 5.3  \\
    \quad w/ PCA & 19 & 1.04 & 0.28 & 0.47 & 0.46 & 5.5 & 0.0  \\
    \quad w/ UMAP & 12 & 1.13  &  0.29  & 0.70 & 0.50 & 25.0 & 8.0  \\
    \textit{Encoder-Encoder} \\ 
    \quad No dim. reduction & 19 & 6.93  &  0.01  & 0.41 & 0.46 & 21.1 & 15.8 \\
    \quad w/ PCA  & 19 & 0.49  &  \textbf{0.54}  & 0.53 & 0.43 & 15.0 & 7.7 \\
    \quad w/ UMAP & 19 & \textbf{0.49}  &  0.53  & 0.51 & 0.48 & \textbf{36.8} & \textbf{21.1} \\
    \textit{Classifier Model} \\
    \quad No dim. reduction & 5 & 1.98  &  0.06  & 0.49 & 0.44 & 0.0 & 0.0 \\
    \quad w/ PCA  & 13 & 0.84  &  0.36  & 0.44 & 0.44  & 7.4 & 0.0  \\
    \quad w/ UMAP & 18 & 0.55  &  0.49  & 0.44 & 0.49 & 5.5 & 5.5  \\
    \textit{Pretrained Decoder} \\
    \quad No dim. reduction & 19 & 2.76  &  0.06  & 0.39 & 0.42  & 16.0 & 11.1 \\
    \quad w/ PCA  & 18 & 1.89  &  0.12  & 0.44 & 0.61 & 5.6 & 5.6  \\
    \quad w/ UMAP & 19 & 1.01  &  0.34  & 0.36 & 0.42 & 11.0 & 11.0  \\
    \multicolumn{3}{@{}l}{\textit{Pretrained Encoder-Decoder}}\\
    \quad No dim. reduction & 5 & 1.62  &  0.16  & 0.44 & 0.48  & 0.0 & 0.0  \\
    \quad w/ PCA  & 8 & 1.74  &  0.20  & 0.37 & 0.46 & 0.0 & 0.0  \\
    \quad w/ UMAP & 5 & 0.75  &  0.44  & 0.46 & 0.46 & 0.0 & 0.0  \\
    \bottomrule 
    
\end{tabularx}
\caption{Internal and external validation metrics for the unsupervised component with the \textbf{K-Means clustering algorithm on the MBIC dataset}.
Internal validation metrics explain intra-cluster separation through higher Silhouette and lower Davies-Bouldin (DB Index) scores. External validity indicates the potential capturing of a voice, measured by the average Purity score and \% of prototypical clusters.}
\label{tab:clusters_mbic}
\end{table*}

\section{Unsupervised Component}\label{sec:dimensionality_reduction_techniques}

\paragraph{Dimensionality Reduction} We perform dimensionality reduction on the hidden states before clustering as follows: a baseline without dimensionality reduction, Principal Component Analysis (PCA; a linear combination of components) and Uniform Manifold Approximation and Projection for Dimension Reduction (UMAP; a non-linear transformation algorithm; \citealp{mcinnes2018umap}). Both PCA \citep{sia2020tired, gupta2019improving} and UMAP \citep{cai2020isotropy, ait2023anisotropy, george2023integrated} 
improve feature representation in high-dimensional latent spaces leading to improved clustering. 

\paragraph{Clustering Algorithms} We used three clustering techniques: K-means \citep{macqueen1967some, scikit-learn}, Gaussian Mixture Models \citep[GMM;][]{rasmussen1999infinite}, and HDBSCAN \citep{mcinnes2017hdbscan}. Each of these techniques have been used to cluster features when paired with either PCA \citep{hosseini2022deep, liu2021primal, asyaky2021improving}, or UMAP \citep{allaoui2020considerably, asyaky2021improving}.

\paragraph{Metrics} We use two \textbf{internal validation} metrics to assess average similarity scores between clusters, namely \textit{Silhouette} \citep{rousseeuw1987silhouettes, scikit-learn} and \textit{Davies-Bouldin Index} \citep{davies1979cluster, scikit-learn}. Silhouette assesses intra-cluster separation and is bound between -1 and 1, with 1 being the best possible score, with a threshold of 0.5 for moderate clusters \citep{shahapure2020cluster, lengyel2019silhouette}. The Davies-Bouldin Index measures intra-cluster dissimilarity, with 0 indicating the lowest possible score \citep{idrus2022distance, karkkainen2000minimization}.

We use \textit{Purity} to assess the \textbf{external validity} of clusters. Purity measures the internal consistency of assigned labels within a cluster and evaluates whether a cluster is prototypical (i.e., representative) across provided labels within a dataset \citep{christodoulopoulos2010two}. In our case, we report both average purity and the percentage of prototypical clusters per method. We define a cluster as prototypical if its metadata label purity (i.e., political leaning and education level) is significantly different from the original dataset metadata label distribution with a threshold of $\pm$ 10\%. These metrics allow us to automatically assess whether a cluster emerging from annotator behaviours during training is linked to any of the annotator labels (e.g., a cluster with high right-leaning metadata label purity) and thus is indicative of a distinct voice.

\begin{table*}[htb]
\centering
\scriptsize
\begin{tabularx}{\textwidth}{@{}cXc p{18mm}@{}}
\toprule
Dataset/Cluster No. & Examples & Bias Label & Distribution\\
\midrule
\multirow{3}[7]{*}{MBIC -1} & British Olympic swimmer Sharron Davies also slammed the concept of transgender athletes. & \cmark & \multirow{3}[5]{*}[2pt]{\includegraphics[width=18mm]{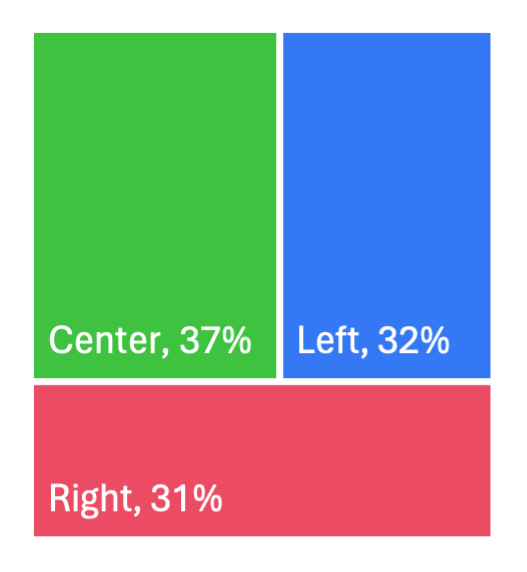}} \\
\cmidrule[0.5pt](lr){2-2}
& BBC Presenter Gabby Logan has said that it is not fair that transgender women can compete in sport alongside biologically female women. & \cmark\\ 
\cmidrule[0.5pt](lr){2-2}
& BBC Presenter Gabby Logan has said that it is not fair that transgender women can compete in sport alongside biologically female women. & \xmark \\ 
\midrule[0.5pt]
\multirow{3}[18]{*}{MBIC -7} & Trump — who has been criticized for painting an overly rosy picture of the outbreak, often contradicting his own health officials - insisted on Friday that his administration was “magnificently organized” and “totally prepared" to address the virus. & \cmark & \multirow{3}[18]{*}{\includegraphics[width=18mm]{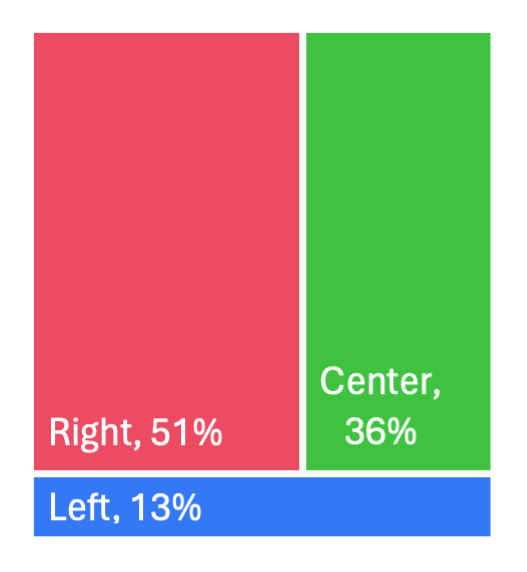}}  \\
\cmidrule[0.5pt](lr){2-2}
&Google declined to offer details beyond Huntley’s tweets, but the unusually public attribution is a sign of how sensitive Americans have become to digital espionage efforts aimed at political campaigns. & \xmark \\ 
\cmidrule[0.5pt](lr){2-2}
&At least 25 transgender or gender-nonconforming people were killed in violent attacks in the United States last year, according to the Human Rights Campaign, which has been tracking anti-trans violence since at least 2015. & \cmark \\ 
\midrule[0.5pt]
& Though conservatives try to demonize Ocasio-Cortez an Omar, their actual policy views are perfectly mainstream. The New York lawmaker proposed a 70 percent tax on top incomes — a view backed by public opinion and many well-respected economists. & \xmark &\multirow{3}[15]{*}{\includegraphics[width=18mm]{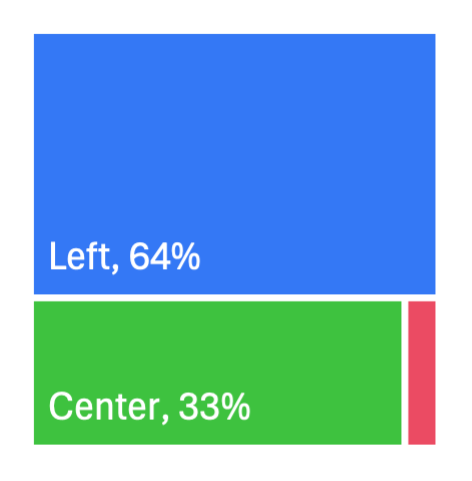}} \\
\cmidrule[0.5pt](lr){2-2}
MBIC -8&British Olympic swimmer Sharron Davies also slammed the concept of transgender athletes. & \xmark \\ 
\cmidrule[0.5pt](lr){2-2}
& At least 25 transgender or gender-nonconforming people were killed in violent attacks in the United States last year, according to the Human Rights Campaign, which has been tracking anti-trans violence since at least 2015. & \cmark \\ 
\bottomrule
\end{tabularx}
\caption{Analysis of clusters on the MBIC dataset with the Encoder-Encoder architecture and UMAP dimensionality reduction. We report the cluster number, representative examples of the cluster, and their paired annotation (\cmark\; for perceived bias, \xmark for no perceived bias). We also show the distribution of annotator characteristics which is indicative of the prototypical nature of each cluster.}
\label{tab:qual-table_mbic}
\end{table*}

\subsection{Results}\label{sec:quantitative_validation} Optimal cluster numbers were automatically calculated using hyperparameter sweeps to maximise the Silhouette score (see \cref{sec:training-details} for more information). \cref{tab:clusters_mbic} shows the clustering of our best performance combination, K-means with a UMAP dimensionality reduction on the MBIC dataset as other configurations performed less optimally as seen in \cref{sec:cluster_metrics}.

\paragraph{Internal Validity Metrics} Overall, dimensionality reduction significantly impacted the quality of the resulting clusters; UMAP consistently outperformed PCA throughout, while no dimensionality reduction showed the worst overall results (for averages, see \cref{sec:dim-averages}). The only exception was Encoder-Encoder, where PCA and UMAP perform comparably. 

Encoder-Encoder performed best overall: being the only model with Silhouette and Davies-Boulding Index scores above/below the respective cutoff points of 0.5, indicating adequate intra-cluster separation for both metrics \citep{shahapure2020cluster, lengyel2019silhouette, idrus2022distance}. Interestingly, the Classifier Model also performed relatively well despite being the lowest-performing of the supervised component.

\paragraph{External Validity Metrics}\label{sec:ext_val} 
Average purity scores are largely inconclusive, as higher scores are not always linked with better performance as indicated by any other evaluative metric. For example, Unpooled Cross Attention with no dimensionality reduction, scores poorly on internal validation metrics, while average purity is the highest across both metadata labels.

Overall, these findings echo those seen in \cref{tab:model_performance}, where models with the lowest APCS scores also had the best performance in internal and external validation metrics. The best-performing model was Encoder-Encoder with UMAP outperforming PCA, followed by Unpooled Cross Attention. 
While UMAP only marginally outperformed PCA in terms of internal validation scores, the label distributions in the clusters resulting from PCA were minimally different when compared to label distributions present in the original data. Finally, we found that prototypical clustering percentage was a strong indicator of capturing representative clusters of voices.

Manual inspection of PCA-formed clusters indicated that clusters formation was mostly based around the most salient features discovered during training, namely the unique annotator tokens or the inter-sentence similarities. A possible reason for this phenomenon could be that PCA reduces dimensionalities to the most salient principal components, which are not conducive to clustering based on contextual features in large language models \citep{cai2020isotropy}. Interestingly, this phenomenon was reproduced with UMAP when instructing the model to focus on finding clusters based on local and not overarching features \citep{mcinnes2018umap}.\footnote{A possible solution to this issue is to remove the top principal components resulting in more salient representations, and thus improve clustering performance \citep{mu2017all}; we leave this for future work.}

\begin{table*}[htb]
\centering
\scriptsize
\renewcommand{\arraystretch}{1.1}
\begin{tabularx}{\textwidth}{@{}cXc p{18mm}@{}}
\toprule
Dataset/Cluster No. & Examples & Agreement Label & Distribution \\
\midrule
\multirow{3}[7]{*}{GWSD -9}& The early 21st-century drought that afflicted Central Asia is the worst in Mongolia in more than 1,000 years, and made harsher by the higher temperatures consistent with man-made global warming. & \cmark & \multirow{3}[7]{*}{\includegraphics[width=18mm]{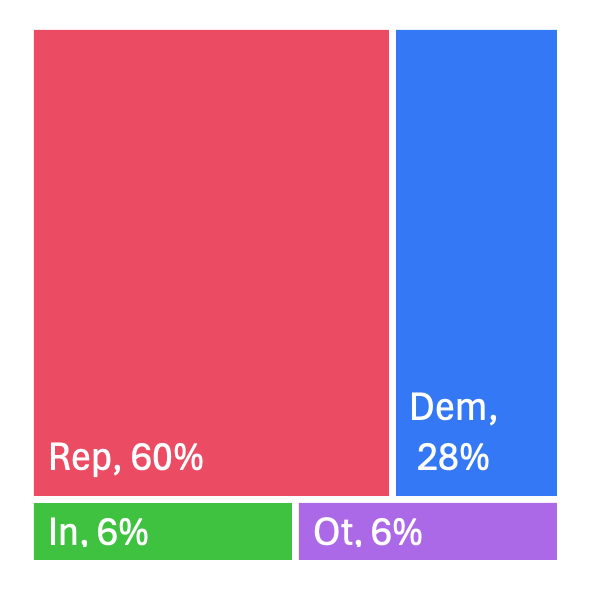}}  \\
\cmidrule[0.5pt](lr){2-2}
& Climate change means the end of shopping. & $\sim$ \\ 
\cmidrule[0.5pt](lr){2-2}
& The oil sands are responsible for just 0.001 percent of global greenhouse emissions & $\sim$ \\ 
\midrule[0.5pt]
\multirow{3}[7]{*}{GWSD -2}& There is a connection between human activity and an assumptive change in global climate. & \cmark & \multirow{3}[7]{*}[3pt]{\includegraphics[width=18mm]{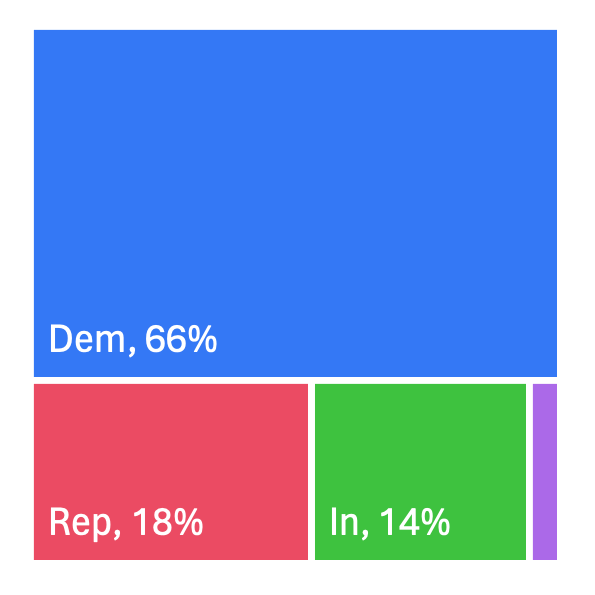}}   \\
\cmidrule[0.5pt](lr){2-2}
&Hiring a White House "climate change czar" would be a good idea. & \cmark \\ 
\cmidrule[0.5pt](lr){2-2}
&Scaring young people into believing that climate change is going to kill young people is child abuse. & \xmark \\ 
\midrule[0.5pt]
\multirow{3}[9]{*}{GWSD -5} & The oil sands are responsible for just 0.001 percent of global greenhouse emissions & \cmark & \multirow{3}[9]{*}{\includegraphics[width=18mm]{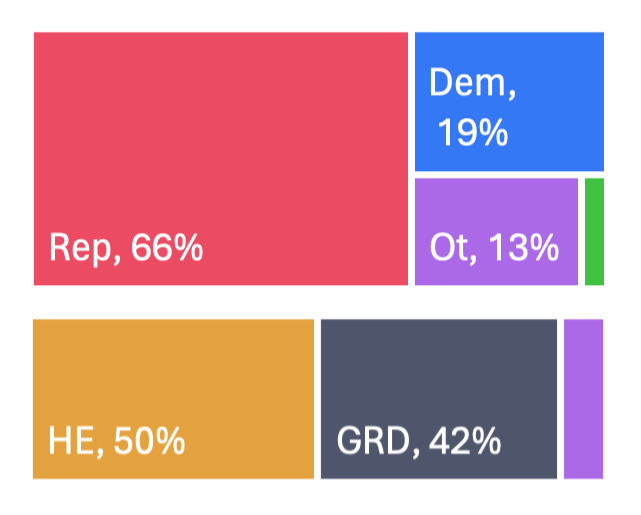}} \\
\cmidrule[0.5pt](lr){2-2}
&This could mean that current I.P.C.C. model predictions for the next century are wrong, and there will be no cooling in the North Atlantic to partially offset the effects of global climate change over North America and Europe. & \cmark \\ 
\cmidrule[0.5pt](lr){2-2}
& Eco-towns could provide an inspiring blueprint for low-carbon living & \xmark \\ 
\bottomrule
\end{tabularx}
\caption{Analysis of clusters on the GWSD dataset using same parameters as the MBIC dataset, and results are shown in a similar fashion (\cmark agree with the statement, \xmark for disagree and $\sim$ for neutral). Distribution of annotator characteristics is provided.}
\label{tab:qual-table_gwsd}
\end{table*}

\section{Qualitative case study}\label{sec:qualitative_validation}

While encouraging, our findings cannot be simply explained through either internal or external validation metrics; to assess whether a cluster is truly indicative of a voice, we looked at the content of the clusters themselves. High label purity of a cluster should be reflected in the text-annotation pair content (i.e., high left-leaning purity should be paired with left-leaning opinions). Given our labels, this can result in three distinct types of voices: majority, minority and inter-minority. 

Majority voice clusters consist of high purity of a majority metadata label (e.g., left-leaning opinions in a left-leaning majority dataset), while minority voices are the same for dataset minority labels (e.g., right-leaning opinions in a left-leaning majority dataset), and inter-minority voices, which are clusters that consist of high purity across combination of metadata labels (e.g., high purity in both right-leaning and highly educated metadata labels in a dataset with left-leaning and non-highly educated majority metadata labels.

To extract our clusters, we used the best-performing combination, i.e., Encoder-Encoder with UMAP and K-means clustering. We pick three prototypical clusters out of a single clustering run, each representing a distinct voice, and discuss them in \cref{tab:qual-table_mbic} and \cref{tab:qual-table_gwsd}.

\subsection{MBIC Dataset} \label{sec:MBIC_results}

\paragraph{MBIC-1 / Minority Voice} This cluster is a prototypical example of minority-led consensus amongst annotators. The cluster's distribution is more even, following the original label distribution closer (44.3\%, 29.1\%, 26.7\% for left, center, and right political lean). Such clusters often contain different annotations for the same sentences, while there is no strong emerging effect from collected labels. 

\paragraph{MBIC-7 / Minority voice} This is a minority voice, with the distribution of labels indicating that the cluster is primarily formed of right-leaning opinions. While Item 1 is expectantly labelled as `bias', Item 3 contains no obvious biased words, despite coming from an obvious place of concern for a marginalised minority.

\paragraph{MBIC-8 / Majority voice} This is an example of a majority dominant cluster. Such clusters are populated by the opinion of the original dataset's distributional majority label although with a much heavier skew, indicating a stable and consistent behaviour of the group. The labelling distribution of this cluster is expected to be populated by left-leaning views and indeed sentences that were previously labelled as biased in non-left-leaning clusters (Item 1 of Cluster 1, and Item 3 of Cluster 7), were consistently found not to be labelled as such.

\subsection{GWSD Dataset}

\paragraph{GWSD-9 / Minority voice} This is an example of a minority cluster, as indicated by the differences in the distribution of the minority label between the cluster and the original data (21\% in the original data, 60\% representation in this cluster). While the expressed opinions within were generally agreeable about climate-changing effects, there was no agreement with more politically charged statements.

\paragraph{GWSD-2 / Majority voice} This is a majority-dominant cluster. Opinions that could be perceived as more political were found to be more common (Item 2), while there was also evidence of general agreement with some strongly politically charged examples (Item 3).

\paragraph{GWSD-5 / Minority-Minority voice} An example of a minority within a minority perspective. Opinions are over-represented by two minority labels, the ``republican'' in terms of political affiliation, and that of the ``higher degree'' in terms of education level (8.4\% label representation in the original dataset). Opinions showed fewer ``neutral'' responses and were generally indicative of a well-informed audience, explicitly agreeing with more technical items such as Item 2 and especially Item 1, which received mostly ``neutral'' scores in other clusters (e.g., Cluster 9).

\section{Conclusion}

We propose a novel framework to identify underlying minority perspectives in data. We compared six distinct model architectures trained on a classification task, without providing any annotator metadata to avoid biasing their training. Subsequently, final hidden states were passed through various methods of dimensionality reduction (UMAP and PCA), with the resulting embeddings used to create clusters through various unsupervised algorithms (K-means, GMM, and HDBSCAN).

The resulting clusters were adequately separated according to internal and external validation metrics. Further qualitative analysis of clusters produced by our best-performing model showcased the ability of our framework to capture perspectives as shown by three distinct types: clusters representative of a minority, a majority, and clusters that captured multiple minority labels, i.e., a minority within a minority.

\section*{Limitations \& Ethical Considerations}

\paragraph{Internal \& External Validity Related}

As shown in \cref{tab:clusters_mbic,sec:cluster_metrics} while internal validation scores \textit{can} be indicative of well-defined clusters of minority perspectives, they are not necessarily so. We explained in \cref{sec:cluster_metrics}, this might be due to our training on unique annotator tokens, which might hinder organic clustering based on behaviour, by providing an alternative and easier to learn signal in unique annotator tokens. 

We aim to expand upon this in future work, by modifying training of our supervised component to incorporate aspects more representative of group behaviours such as inter and intra annotator disagreement \citep{abercrombie2023consistency, leonardelli2023semeval, uma2021learning, uma-etal-2021-semeval}. This would expand upon limitations of disagreement-bases approaches described in \cref{sec:related-work}, by enabling group behavioural signals, as indicated by annotator agreement / disagreement, to be captured on the dataset-level. Furthermore, incorporation of such methodologies into our framework would further expand upon the limitations of disagreement-based methodologies by allowing for any number of voices to be expressed.

\paragraph*{Automatic Detection of Voices}\label{sec:lim_autodetect}

A current limitation of the framework is the ability to automatically assess the performance of each combination without manual inspection. While necessary at this step to prove the efficacy of our framework, we aim to expand this in future work by introducing a a component that automatically extracts information from each cluster to allow for identification of voice without the need of matching clusters with metadata labels post hoc.

We aim to employ a similar methodology to \citet{fleisig2023majority}, whose pipeline includes a GPT-2 based component that predicts the demographic group targeted by a given text. We aim to include similar components to extrapolate attitudinal and behavioural indicators of formed clusters via analysing the text-annotation pairs to generate labels representative of each captured voice similarly to how research in sentiment analysis, has previously classified opinions on politically charged data \citep{dorle2018political, kazienko2023human, ansari2020analysis}.

\paragraph*{Labels and further marginalisation of minorities}

Our model uses labels procured during data gathering to validate emergent clusters. However, the labelling gathering process can potentially be an erasing process towards minorities in and of itself \citep{hovy2021five, chandrabose2021overview}. For example, the labelling process can discriminate against socially marginalised minorities by not providing options consistent with an individual's identity \citep{chandrabose2021overview, jo2020lessons}. 

In our case, we encountered this limitation with the GWSD dataset \citep{luo2020detecting}, which collected categorical labels about political affiliation of participants. Beyond the three primary labels ("Democrat", "Independent", "Republican"), the rest were aggregated into the "other" label. This resulted in a minority so small that our clustering methodology could not adequately disentangle it from the rest. Directions aimed towards future research as explained in \cref{sec:lim_autodetect} should address these concerns for future iterations of our framework.

\paragraph*{Dual Use of the Model}

An unfortunate outcome of methodologies aim to capture and expressed more nuanced perspectives can lead to identification of marginalised minority perspectives in datasets, which can lead to concerning practice of their removal in order to enhance a model's general performance \citep{xu-etal-2021-detoxifying, sun-etal-2019-mitigating}. Nevertheless, \citet{gaci2023targeting} has also proposed that methodologies that identify minority perspectives can be used to curate datasets in order to amplify voices of specific marginalised groups. 

We urge researchers to be transparent in their indented use of our framework, and to follow ethical frameworks and solutions that have been previously highlighted by the field in from the data collection process to model training and intended use \citep{hovy2021five,blodgett-etal-2020-language, navigli2023biases, leidner2017ethical, shmueli2021beyond}.

\bibliography{anthology,custom}

\appendix
\clearpage
\section{Training Details}\label{sec:training-details}

To aid in reproducibility, we report all training details and any relevant hyperparameters. 

\subsection{Hyperparameters}

All models were trained using a single NVIDIA A40 GPU. A total of 1080 hours were used during training of all models. For all models, we used the AdamW optimizer \citep{loshchilov2017decoupled} during training with weight decay \num{0.01}. We report hyperparameters for each model and dataset in \cref{tab:hyperparam-training}.From small performance gains during preliminary experiments, we disable bias across all linear layers.

Cluster training hyperparameters can be found in \cref{tab:hyperparam-cluster}. Across every model, we found that when comparing hyperparameters for both PCA and UMAP converged to the same choices. For both methods, we found that 2 components yielded the best results. Additionally, for UMAP, we found that the optimal number of neighbours were found to be between 80--100 across all models,with a minimum distance ranging from 0.8 to 1 to yield better clustering performance. 

\begin{table}[!hb]
\centering
\footnotesize
\renewcommand{\arraystretch}{1.3}
\begin{tabularx}{\linewidth}{Xc@{}}
    \toprule
    Hyperparameter & Value \\
    \midrule
    \multicolumn{2}{@{}l}{\textit{Unpooled Cross Attention}} \\
    Model name & \verb|google/t5-v1_1-large| \\
    Downsampling n. of layers & 0-3 \\
    N. warmup steps & 0- 800 \\
    Learning rate & 0.0001 - 1e-08 \\
    \multicolumn{2}{@{}l}{Pooled Cross Attention} \\
    Model name & \verb|google/t5-v1_1-large| \\
    Ann dim. factor & 1-6 \\
    Downsampling n. of layers & 0-3 \\
    N. warmup steps & 0- 800 \\
    Learning rate &0.0001 - 1e-08 \\
    \multicolumn{2}{@{}l}{Encoder-Encoder} \\
    Model name & \verb|google/t5-v1_1-large| \\
    Downsampling n. of layers & 0-3 \\
    N. warmup steps & 0- 800 \\
    Learning rate & 0.0001 - 1e-08 \\
    \multicolumn{2}{@{}l}{Classifier Model} \\
    Model name & \verb|roberta-large| \\ 
    N. warmup steps & 0- 800 \\
    Learning rate & 1e-11 - 1e-3 \\
    \multicolumn{2}{@{}l}{Pretrained Decoder} \\
    Model name & \verb|gpt2-large| \\
    Downsampling n. of layers & 0-3 \\
    N. warmup steps & 0- 800 \\
    Learning rate & 0.0001 - 1e-08 \\
    \multicolumn{2}{@{}l}{Pretrained Encoder-Decoder} \\
    Model name & \verb|google/t5-v1_1-large| \\
    Downsampling n. of layers & 0-3 \\
    N. warmup steps & 0- 800 \\
    Learning rate & 0.0001 - 1e-08 \\
    \bottomrule
\end{tabularx}
\caption{Hyperparameters for all supervised models on each of our chosen datasets, obtained from running running a hyperparameter sweep for 12 hours.}
\label{tab:hyperparam-training}
\end{table}

\begin{table}[!hb]
\centering
\footnotesize
\renewcommand{\arraystretch}{1.3}
\begin{tabularx}{\linewidth}{Xc@{}}
    \toprule
    Hyperparameter & Value \\
    \midrule
    \multicolumn{2}{@{}l}{\textit{PCA}} \\
    Cluster ranges & 2 - 19 \\
    N components & 2-40 \\
    \multicolumn{2}{@{}l}{GMM} \\
    Cluster ranges & 2-19 \\
    \multicolumn{2}{@{}l}{HDBSCAN} \\
    Eps & 0.0 - 1.0 \\
    Min samples & 2 - 100 \\
    Min cluster size & 2 - 100 \\
    \bottomrule
\end{tabularx}
\caption{Hyperparameters for all clustering methods on each of our chosen datasets, obtained from running running a hyperparameter sweep for 12 hours.}
\label{tab:hyperparam-cluster}
\end{table}

\subsection{Dimensionality Reduction}\label{sec:dim-averages}

We report internal validity evaluation score averages across dimensionality reduction techniques in \cref{fig:dimensionality_reduction}.

\begin{table}[!ht]
    \centering
    \footnotesize
    \sisetup{mode=match,tight-spacing=true,table-format=1.3}
    \renewcommand{\arraystretch}{1.3}
    \begin{tabularx}{\linewidth}{@{} X SS @{}}
        \toprule
        & {Davies-Bouldin Index} & {Silhouette} \\
        \midrule
        No dim. reduction & 3.655 & 0.073 \\
        w/ PCA & 0.491 & 0.56 \\
        w/ UMAP & 0.565 & 0.53 \\
        \bottomrule
    \end{tabularx}
    \caption{Dimensionality reduction effect on internal validity scores}
    \label{fig:dimensionality_reduction}
\end{table}

\section{Visual Representation of Models used in Training Component}\label{sec:model_image}
\begin{figure*}[bt!]
    \centering
    \includegraphics[width=\textwidth]{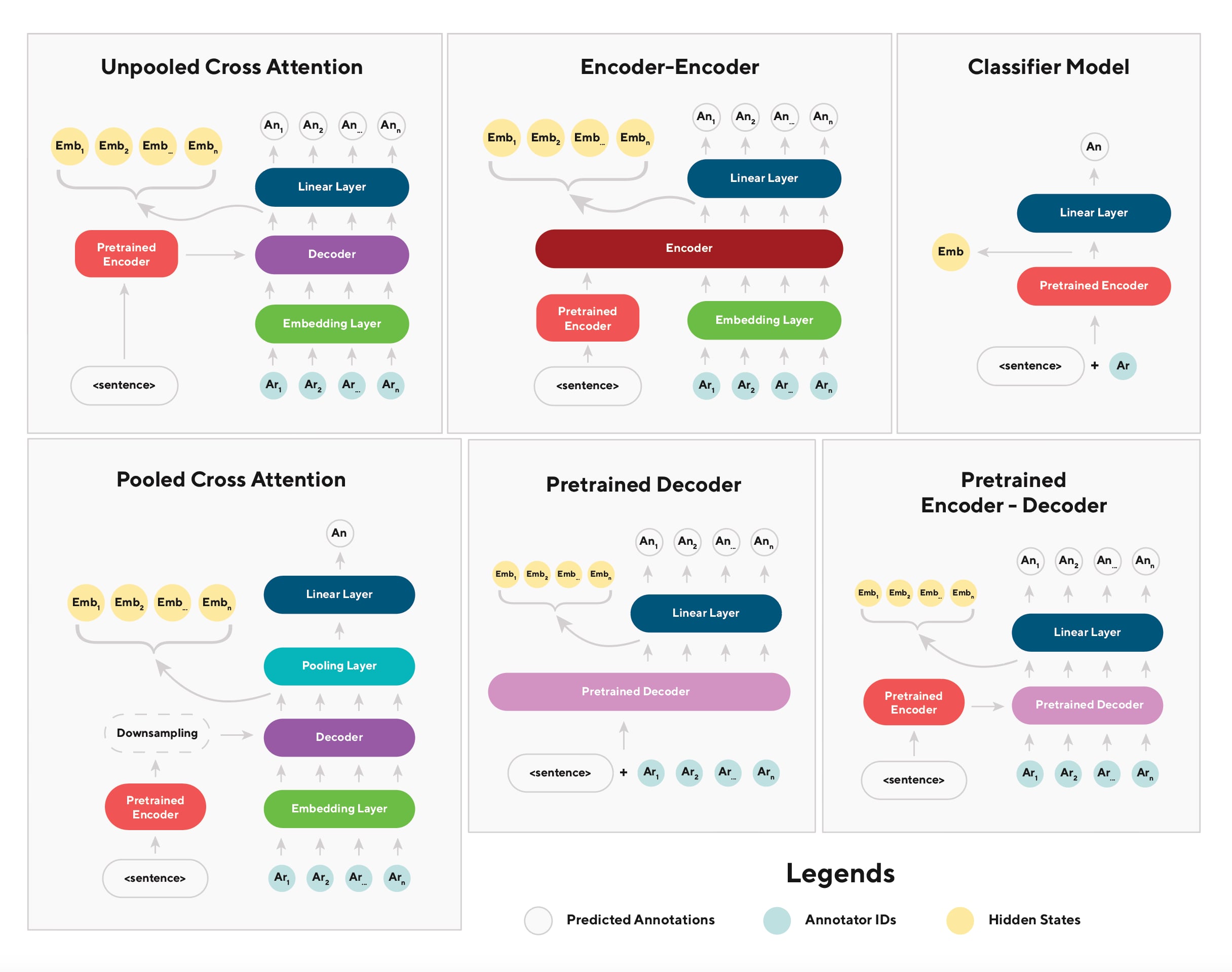}     
    \caption{Training component: 6 modelling architectures for extracting hidden states (denoted with a yellow circle as $Emb_n$) used as input for the Clustering component.}
    \label{fig:model}
 \end{figure*}

Visual depictions of all model architectures seen in \cref{fig:model}.

\section{Cluster Metrics}\label{sec:cluster_metrics}

\subsection{GWSD Cluster Validity Scores - Kmeans}\label{sec:gwsd-kmeans-validity-section}

We report the GWSD internal and external validation metrics resulting from our clustering using a k-means algorithm and our various employed dimensionality reduction techniques in \cref{tab:clusters_gwsd_kmeans}.

\begin{table*}[!ht]
\centering
\footnotesize
\sisetup{mode=match,tight-spacing=true,separate-uncertainty=true,table-align-uncertainty=true,group-separator=\pm,detect-weight=true,detect-inline-weight=math}
\renewcommand{\arraystretch}{1.3}
\setlength{\tabcolsep}{6pt}
\begin{tabular}{@{} l S[table-format=2] *{4}{S[table-format=1.2]} *{2}{S[table-format=2.1]} @{}}
    \toprule
    &&&&
    \multicolumn{2}{c}{\textit{Purity} $\uparrow$} &
    \multicolumn{2}{c}{\textit{Prototypical cluster \%} $\uparrow$}
    \\
    & {\# Clusters} 
    & {DB Index $\downarrow$} 
    & {Silhouette $\uparrow$} 
    & {Political} 
    & {Education} 
    & {Political} 
    & {Education} 
    \\
    \midrule
    \multicolumn{1}{@{}l}{\textit{GWSD - Kmeans}} \\
    Cross Attention &  \\
    \quad No dim. reduction  & 19 & 1.95  &  0.17  & 0.46 & 0.43  & 0.00  & 0.05\\\
    \quad w/ PCA  & 17 & 0.45  &  0.61  & 0.53 & 0.53  & 0.00  & 0.00\\
    \quad w/ UMAP & 18 & 1.05  &   \textbf{0.49}  & 0.51 & 0.44  & 0.22  & 0.00\\
    Pooled Cross Attention &  \\ 
    \quad No dim. reduction & 16 & 2.76  &  0.07  & 0.43 & 0.44  & 0.06  & 0.00\\
    \quad w/ PCA  & 19 & 0.79  &  0.38  & 0.43 & 0.47  & 0.16  & 0.00\\
    \quad w/ UMAP & 19 & \textbf{0.47}  &  0.55  & 0.49 & 0.40  & 0.05  & 0.11\\
    Encoder-Encoder &  \\ 
    \quad No dim. reduction & 18 & 5.77  &  0.02  & 0.53 & 0.34  & 0.28  & \bfseries 0.33\\
    \quad w/ PCA  & 19 & 0.84  &  0.34  & 0.40 & 0.60  & 0.11  & 0.00\\
    \quad w/ UMAP & 15 & 0.50  &  0.54  & \textbf{0.69} & 0.54  & \bfseries 0.40  & 0.20\\
    Classifier Model & \\
    \quad No dim. reduction & 19 & 1.95  &  0.17  & 0.46 & 0.43  & 0.00  & 0.05\\
    \quad w/ PCA  & 17 & 0.45  &  0.61  & 0.53 & 0.53  & 0.00  & 0.00\\
    \quad w/ UMAP  & 18 & 1.05  & \textbf{0.49}  & 0.51 & 0.44  & 0.22  & 0.00\\
    Pretrained Decoder & \\
    \quad No dim. reduction & 19 & 2.83  &  0.09  & 0.61 & 0.47  & 0.11  & 0.05\\
    \quad w/ PCA  & 19 & 0.47  &  0.59  & 0.42 & 0.44  & 0.16  & 0.00\\
    \quad w/ UMAP & 17 & 0.52  &  0.53  & 0.51 & \textbf{0.58}  & 0.00  & 0.00\\
    Pretrained Encoder-Decoder & \\
    \quad No dim. reduction & 19 & 2.53  &  0.06  & 0.48 & 0.55  & 0.05  & 0.05\\
    \quad w/ PCA & 19 & 0.83  &  0.34  & 0.45 & 0.52  & 0.11  & 0.11\\
    \quad w/ UMAP & 17 & 0.84  &  0.34  & 0.36 & 0.57  & 0.00  & 0.06\\
    \bottomrule
    
\end{tabular}
\caption{Internal and external validation metrics for the K-means clustering technique on the GWSD dataset. Internal validation metrics explain intra-cluster separation through higher Silhouette and lower Davies-Bouldin (DB Index) scores. External validity, which indicates the potential of having captured a voice, is measured via the average Purity score and \% of prototypical clusters.}\label{tab:clusters_gwsd_kmeans}
\end{table*}

\subsection{GWSD Cluster Validity Scores - GMM}\label{sec:gwsd-gmm-validity-section}

We report the GWSD internal and external validation metrics resulting from our clustering using a GMM algorithm and our various employed dimensionality reduction techniques in \cref{tab:clusters_gwsd_gmm}. This methodology resulted in cluster metrics which were not as optimal as those of the K-means solutions.

\begin{table*}[t]
\centering
\footnotesize
\sisetup{mode=match,tight-spacing=true,separate-uncertainty=true,table-align-uncertainty=true,group-separator=\pm,detect-weight=true,detect-inline-weight=math}
\renewcommand{\arraystretch}{1.3}
\setlength{\tabcolsep}{6pt}
\begin{tabular}{@{} l S[table-format=2] *{4}{S[table-format=1.2]} *{2}{S[table-format=2.1]} @{}}
    \toprule
    &&&&
    \multicolumn{2}{c}{\textit{Purity} $\uparrow$} &
    \multicolumn{2}{c}{\textit{Prototypical cluster \%} $\uparrow$}
    \\
    & {\# Clusters} 
    & {DB Index $\downarrow$} 
    & {Silhouette $\uparrow$} 
    & {Political} 
    & {Education} 
    & {Political} 
    & {Education} 
    \\
    \midrule
    \multicolumn{1}{@{}l}{\textit{GWSD -GMM}} \\
    Unpooled Cross Attention &  \\
    \quad No dim. reduction & 5 & 12.54  &  0.00  & 0.44 & 0.55  & 0.00  & 0.00 \\
    \quad w/ PCA  & 5 & 8.13  &  0.00  & 0.44 & 0.55  & 0.00  & 0.00 \\
    \quad w/ UMAP & 5 & 8.02  &  0.01  & 0.44 & 0.55  & 0.00  & 0.00 \\
    Pooled Cross Attention &  \\ 
    \quad No dim. reduction & 6 & 3.73  &  0.04  & 0.46 & 0.57  & 0.00  & 0.00 \\
    \quad w/ PCA  & 6 & 2.68  &  0.05  & \textbf{0.46} & \textbf{0.57}  & 0.00  & 0.00 \\
    \quad w/ UMAP & 7 & 2.31  &  0.08  & 0.37 & 0.46  & 0.00  & 0.00 \\
    Encoder-Encoder &  \\ 
    \quad No dim. reduction & 5 & 9.30  &  0.01  & 0.44 & 0.47  & 0.00  & 0.00 \\
    \quad w/ PCA  & 5 & 4.09  &  0.03  & 0.44 & 0.47  & 0.00  & 0.00 \\
    \quad w/ UMAP & 5 & 5.57  &  0.03  & 0.44 & 0.47  & 0.00  & 0.00 \\
    Classifier Model & \\
    \quad No dim. reduction  & 5 & 1.87  &  0.19  & 0.43 & 0.51  & 0.00  & 0.00 \\
    \quad w/ PCA  & 5 & 1.48  &  \textbf{0.33}  & 0.43 & 0.51  & 0.00  & 0.00 \\
    \quad w/ UMAP & 12 & 3.02  &  0.05  & 0.42 & 0.50  & 0.08  & 0.00 \\
    Pretrained Decoder & \\
    \quad No dim. reduction & 19 & 3.12  &  0.05  & 0.41 & 0.50  & 0.05  & 0.00 \\
    \quad w/ PCA  & 6 & 1.72  &  0.18  & 0.44 & 0.48  & 0.00  & 0.00 \\
    \quad w/ UMAP & 5 & \textbf{1.75}  &  0.20  & 0.47 & 0.53  & 0.00  & 0.00\ \\
    Pretrained Encoder-Decoder & \\
    \quad No dim. reduction & 5 & 3.39  &  0.05  & 0.47 & 0.48  & 0.00  & 0.00 \\
    \quad w/ PCA  & 6 & 2.90  &  0.00  & 0.44 & 0.56  & 0.00  & 0.00 \\
    \quad w/ UMAP & 11 & 2.51  &  0.06  & 0.45 & 0.43  & \bfseries 0.09 & 0.00 \\
    \bottomrule
    
\end{tabular}
\caption{Internal and external validation metrics for the GMM clustering technique on the GWSD dataset. Internal validation metrics explain intra-cluster separation through higher Silhouette and lower Davies-Bouldin (DB Index) scores. External validity, which indicates the potential of having captured a voice, is measured via the average Purity score and \% of prototypical clusters.}\label{tab:clusters_gwsd_gmm}
\end{table*}

\subsection{GWSD Cluster Validity Scores - HDBSCAN}\label{sec:gwsd-hdbscan-validity-section}

We report the GWSD internal and external validation metrics resulting from our clustering using an HDBSCAN algorithm and our various employed dimensionality reduction techniques in \cref{tab:clusters_gwsd_hdbscan}. Unfortunately, this methodology resulted in either large cluster numbers too large to be adequately analysed manually, or with metrics not as optimal as those of the K-means solutions.

\begin{table*}[t]
\centering
\footnotesize
\sisetup{mode=match,tight-spacing=true,separate-uncertainty=true,table-align-uncertainty=true,group-separator=\pm,detect-weight=true,detect-inline-weight=math}
\renewcommand{\arraystretch}{1.3}
\setlength{\tabcolsep}{6pt}
\begin{tabular}{@{} l S[table-format=2] *{4}{S[table-format=1.2]} *{2}{S[table-format=2.1]} @{}}
    \toprule
    &&&&
    \multicolumn{2}{c}{\textit{Purity} $\uparrow$} &
    \multicolumn{2}{c}{\textit{Prototypical cluster \%} $\uparrow$}
    \\
    & {\# Clusters} 
    & {DB Index $\downarrow$} 
    & {Silhouette $\uparrow$} 
    & {Political} 
    & {Education} 
    & {Political} 
    & {Education} 
    \\
    \midrule
    \multicolumn{1}{@{}l}{\textit{GWSD- HDBSCAN}} \\
    Unpooled Cross Attention &  \\
    \quad No dim. reduction & 407 & 0.62  &  0.57  & \textbf{1.00} & \textbf{1.00}  & \bfseries  0.96 & \bfseries 1.00  \\
    \quad w/ PCA  & 4 & 10.10  &  0.05  & 0.50 & 0.50  & 0.25  & 0.50 \\
    \quad w/ UMAP & 3 & 17.35  &  0.01  & 0.57 & 0.57  & 0.33  & 0.33 \\
    Pooled Cross Attention &  \\ 
    \quad No dim. reduction & 191 & 1.25  &  0.30  & 0.80 & 0.60  & 0.59  & 0.35 \\
    \quad w/ PCA  &  3 & 2.47  &  0.01  & 0.60 & 0.50  & 0.33  & 0.00\\
    \quad w/ UMAP & 173 & 0.23  &  0.85  & 0.75 & 0.38  & 0.59  & 0.35 \\
    Encoder-Encoder &  \\ 
    \quad No dim. reduction & 4 & 9.53  &  0.01  & 0.67 & 0.67  & 0.50  & 0.25 \\
    \quad w/ PCA  & 5 & 6.99  &  0.03  & 0.43 & 0.57  & 0.00  & 0.40 \\
    \quad w/ UMAP & 4 & 21.22  &  0.07  & 0.52 & 0.92  & 0.25  & 0.25 \\
    Classifier Model & \\
    \quad No dim. reduction & 211 & 0.14  &  0.95  & 0.50 & 0.62  & 0.59  & 0.35 \\
    \quad w/ PCA  & 210 & \textbf{0.13}  &  0.95  & 0.50 & 0.62  & 0.59  & 0.34 \\
    \quad w/ UMAP & 3 & 3.20  &  0.14  & 0.51 & 0.42  & 0.00  & 0.00 \\
    Pretrained Decoder & \\
    \quad No dim. reduction & 210 & 1.21  &  0.62  & 0.50 & 0.62  & 0.60  & 0.35 \\
    \quad w/ PCA  & 204 & 1.14  &  0.52  & 0.40 & 0.60  & 0.57  & 0.38 \\
    \quad w/ UMAP & 210 & 0.78  &  \textbf{0.98}  & 0.50 & 0.62  & 0.59  & 0.35 \\
    Pretrained Encoder-Decoder & \\
    \quad No dim. reduction & 3 & 0.72  &  0.25  & 0.50 & 0.50  & 0.33  & 0.33 \\
    \quad w/ PCA  & 3 & 2.31  &  0.04  & 0.50 & 0.50  & 0.33  & 0.33\ \\
    \quad w/ UMAP & {---} &{---} &{---} &{---} & {---}&{---} &{---}\\
    \bottomrule
    
\end{tabular}
\caption{Internal and external validation metrics for the HDBSCAN clustering technique on the GWSD dataset. Internal validation metrics explain intra-cluster separation through higher Silhouette and lower Davies-Bouldin (DB Index) scores. External validity, which indicates the potential of having captured a voice, is measured via the average Purity score and \% of prototypical clusters. Missing runs indicate that the cluster number computed was equal to the amount of text-annotation pairs, proving the solution invalid.}\label{tab:clusters_gwsd_hdbscan}
\end{table*}

\subsection{MBIC Cluster Validity Scores- GMM}\label{sec:mbic-gmm-validity-section}

We report the MBIC internal and external validation metrics resulting from our clustering using a GMM algorithm and our various employed dimensionality reduction techniques in \cref{tab:clusters_MBIC_gmm}. Unfortunately, this methodology also resulted in cluster metrics which were not as optimal as those of the K-means solutions.

\begin{table*}[t]
\centering
\footnotesize
\sisetup{mode=match,tight-spacing=true,separate-uncertainty=true,table-align-uncertainty=true,group-separator=\pm}
\renewcommand{\arraystretch}{1.3}
\setlength{\tabcolsep}{6pt}
\begin{tabular}{@{} l S[table-format=3] *{4}{S[table-format=1.2]} *{2}{S[table-format=2.1]} @{}}
    \toprule
    &&&&
    \multicolumn{2}{c}{\textit{Purity} $\uparrow$} &
    \multicolumn{2}{c}{\textit{Prototypical cluster \%} $\uparrow$}
    \\
    & {\# Clusters} 
    & {DB Index $\downarrow$} 
    & {Silhouette $\uparrow$} 
    & {Political} 
    & {Education} 
    & {Political} 
    & {Education} 
    \\
    \midrule
    \multicolumn{1}{@{}l}{\textit{MBIC- GMM}} \\
    Unpooled Cross Attention &  \\
    \quad No dim. reduction & 19 & 7.50  &  0.01  & \textbf{0.66} & 0.54  & \bfseries 0.32 & 0.05 \\
    \quad w/ PCA  & 5 & 8.11  &  0.00  & 0.41 & 0.46  & 0.00  & 0.00 \\
    \quad w/ UMAP & 5 & 8.22  &  0.00  & 0.41 & 0.46  & 0.00  & 0.00 \\
    Pooled Cross Attention &  \\ 
    \quad No dim. reduction & 19 & 4.04  &  0.02  & 0.37 & 0.46  & \bfseries 0.32  & 0.05 \\
    \quad w/ PCA  & 8 & 4.09  &  0.00  & 0.45 & \textbf{0.56}  & 0.12  & 0.00 \\
    \quad w/ UMAP & 5 & 7.83  &  0.01  & 0.45 & 0.51  & 0.00  & 0.00 \\
    Encoder-Encoder &  \\ 
    \quad No dim. reduction & 19 & 8.81  &  0.00  & 0.50 & 0.33  & 0.21  & \bfseries 0.21 \\
    \quad w/ PCA  & 5 & 9.50  &  0.00  & 0.47 & 0.48  & 0.20  & 0.20 \\
    \quad w/ UMAP & 5 & 8.87  &  0.00  & 0.47 & 0.48  & 0.20  & 0.20 \\
    Classifier Model & \\
    \quad No dim. reduction & {---} &{---} &{---} &{---} & {---}&{---} &{---}\\
    \quad w/ PCA  & {---} &{---} &{---} &{---} & {---}&{---} &{---}\\
    \quad w/ UMAP & {---} &{---} &{---} &{---} & {---}&{---} &{---}\\ 
    Pretrained Decoder & \\
    \quad No dim. reduction  & 5 & 3.67  &  0.03  & 0.44 & 0.46  & 0.00  & 0.00 \\
    \quad w/ PCA & 16 & 2.83  &  0.01  & 0.52 & 0.32  & 0.00  & 0.00 \\
    \quad w/ UMAP & 18 & 7.50  &  0.01  & 0.53 & 0.50  & 0.17  & 0.00 \\
    Pretrained Encoder-Decoder & \\
    \quad No dim. reduction & 6 & 1.76  &  0.14  & 0.47 & 0.46  & 0.00  & 0.00 \\
    \quad w/ PCA  & 5 & 2.27  &  0.03  & 0.49 & 0.48  & 0.00  & 0.00 \\
    \quad w/ UMAP & 5 & \textbf{0.58}  &  \textbf{0.43}  & 0.49 & 0.48  & 0.00  & 0.00 \\
    \bottomrule
    
\end{tabular}
\caption{Internal and external validation metrics for the GMM clustering technique on the GWSD dataset. Internal validation metrics explain intra-cluster separation through higher Silhouette and lower Davies-Bouldin (DB Index) scores. External validity, which indicates the potential of having captured a voice, is measured via the average Purity score and \% of prototypical clusters. Rows with missing labels indicate inability of the GMM clustering technique to create a solution within the allotted train time for the respective configuration's hyperparameter sweep.}\label{tab:clusters_MBIC_gmm}
\end{table*}

\subsection{MBIC Cluster Validity Scores- HDBSCAN}\label{sec:mbic-hdbscan-validity-section}

We report the MBIC internal and external validation metrics resulting from our clustering using a HDBSCAN algorithm and our various employed dimensionality reduction techniques in \cref{tab:clusters_MBIC_HDBSCAN}. Unfortunately, this methodology also resulted in either large cluster numbers too large to be adequately analysed manually, or with metrics not as optimal as those of the K-means solutions.

\begin{table*}[t]
\centering
\footnotesize
\sisetup{mode=match,tight-spacing=true,separate-uncertainty=true,table-align-uncertainty=true,group-separator=\pm,detect-weight=true,detect-inline-weight=math}
\renewcommand{\arraystretch}{1.3}
\setlength{\tabcolsep}{6pt}
\begin{tabular}{@{} l S[table-format=3] *{4}{S[table-format=1.2]} *{2}{S[table-format=2.1]} @{}}
    \toprule
    &&&&
    \multicolumn{2}{c}{\textit{Purity} $\uparrow$} &
    \multicolumn{2}{c}{\textit{Prototypical cluster \%} $\uparrow$}
    \\
    & {\# Clusters} 
    & {DB Index $\downarrow$} 
    & {Silhouette $\uparrow$} 
    & {Political} 
    & {Education} 
    & {Political} 
    & {Education} 
    \\
    \midrule
    \multicolumn{1}{@{}l}{\textit{MBIC- HDBSCAN}} \\
    Unpooled Cross Attention &  \\
    \quad No dim. reduction & 862 & 1.01  &  0.71  & \textbf{1.00} & \textbf{1.00}  & \bfseries 1.00  & \bfseries 1.00  \\
    \quad w/ PCA  & 862 & \textbf{0.86}  &  \textbf{0.72}  & \textbf{1.00} & \textbf{1.00}  & \bfseries 1.00  & \bfseries 1.00 \\
    \quad w/ UMAP & 862 & 1.30  &  0.21  & \textbf{1.00} & \textbf{1.00}  & \bfseries 1.00  & \bfseries 1.00 \\
    Pooled Cross Attention &  \\ 
    \quad No dim. reduction & 218 & 1.30  &  0.20  & \textbf{1.00} & \textbf{1.00}  & 0.85  & 0.70 \\
    \quad w/ PCA  & 218 & 1.26  &  0.29  & \textbf{1.00} & \textbf{1.00}  & 0.85  & 0.70 \\
    \quad w/ UMAP & 218 & 2.85  &  0.80  & \textbf{1.00} & \textbf{1.00}  & 0.85  & 0.70 \\
    Encoder-Encoder &  \\ 
    \quad No dim. reduction & 5 & 4.18  &  0.00  & \textbf{1.00} & \textbf{1.00}  & 0.60  & 0.60 \\
    \quad w/ PCA  & 3 & 3.59  &  0.06  & \textbf{1.00} & \textbf{1.00}  & 0.67  & 0.33 \\
    \quad w/ UMAP & 5 & 4.10  &  0.04  & \textbf{1.00} & \textbf{1.00}  & 0.60  & 0.60 \\
    Classifier Model & \\
    \quad No dim. reduction & 3 & 2.70  &  0.15  & 0.50 & 0.58  & 0.00  & 0.00 \\
    \quad w/ PCA  & 3 & 1.81  &  0.04  & 0.67 & 0.67  & 0.33  & 0.33 \\
    \quad w/ UMAP & 3 & 1.93  &  0.56  & 0.46 & 0.55  & 0.00  & 0.00 \\
    Pretrained Decoder & \\
    \quad No dim. reduction & 185 & 1.22  &  0.45  & 0.44 & 0.56  & 0.68  & 0.44 \\
    \quad w/ PCA  & 168 & 2.45  &  0.07  & 0.43 & 0.57  & 0.74  & 0.39 \\
    \quad w/ UMAP & 168 & 1.11  &  0.63  & 0.43 & 0.57  & 0.74  & 0.39 \\
    Pretrained Encoder-Decoder & \\
    \quad No dim. reduction & 3 & 1.27  &  0.19  & 0.50 & 0.50  & 0.33  & 0.33 \\
    \quad w/ PCA  & 3 & 2.78  &  0.04  & 0.53 & 0.47  & 0.33  & 0.00 \\
    \quad w/ UMAP  & 3 & 3.29  &  0.08  & 0.53 & 0.49  & 0.33  & 0.00 \\
    \bottomrule
    
\end{tabular}
\caption{Internal and external validation metrics for the HDBSCAN clustering technique on the MBIC dataset. Internal validation metrics explain intra-cluster separation through higher Silhouette and lower Davies-Bouldin (DB Index) scores. External validity, which indicates the potential of having captured a voice, is measured via the average Purity score and \% of prototypical clusters.}\label{tab:clusters_MBIC_HDBSCAN}
\end{table*}

\end{document}